\pdfoutput=1
\documentclass[11pt]{article}

\usepackage[final]{acl}
 
\usepackage{times}
\usepackage{latexsym}

\usepackage[T1]{fontenc}
\usepackage[utf8]{inputenc}

\usepackage{amsfonts}

\usepackage{microtype}

\usepackage{inconsolata}

\usepackage{graphicx}

\usepackage{amsmath}
\usepackage{booktabs}
\usepackage{multirow}
\usepackage{tcolorbox}
\tcbuselibrary{breakable}

\usepackage{hyperref}
\usepackage{xcolor} 
\title{ChartGaze: Enhancing Chart Understanding in LVLMs\\ with Eye-Tracking Guided Attention Refinement}

\author{
Ali Salamatian$^1$ \quad
Amirhossein Abaskohi$^1$ \quad
Wan-Cyuan Fan$^{1,2}$ \\
{\bf Mir Rayat Imtiaz Hossain}$^{1,2}$ \quad
{\bf Leonid Sigal}$^{1,2,3}$ \quad
{\bf Giuseppe Carenini}$^1$ 
\\ \\
$^1$University of British Columbia ~~~ $^2$Vector Institute for AI ~~~
$^3$CIFAR AI Chair\\
\texttt{alisalam@student.ubc.ca} \\
\texttt{\{aabaskoh, wancyuan, rayat137, lsigal, carenini\}@cs.ubc.ca}
}

\begin{document}
\maketitle 
\begin{abstract}
Charts are a crucial visual medium for communicating and representing information. While Large Vision-Language Models (LVLMs) have made progress on chart question answering (CQA), the task remains challenging, particularly when models attend to irrelevant regions of the chart. In this work, we present ChartGaze, a new eye-tracking dataset that captures human gaze patterns during chart reasoning tasks. Through a systematic comparison of human and model attention, we find that LVLMs often diverge from human gaze, leading to reduced interpretability and accuracy. To address this, we propose a gaze-guided attention refinement that aligns image-text attention with human fixations. Our approach improves both answer accuracy and attention alignment, yielding gains of up to 2.56 percentage points across multiple models. These results demonstrate the promise of incorporating human gaze to enhance both the reasoning quality and interpretability of chart-focused LVLMs\footnote{Code and dataset are publicly available at \href{https://github.com/alisalamatian1/ChartGaze}{ChartGazeCode} and \href{https://huggingface.co/datasets/alisalam/ChartGaze}{ChartGazeData}, respectively.}.
\end{abstract}

\begin{figure}[t]
    \centering
    \includegraphics[width=\linewidth]{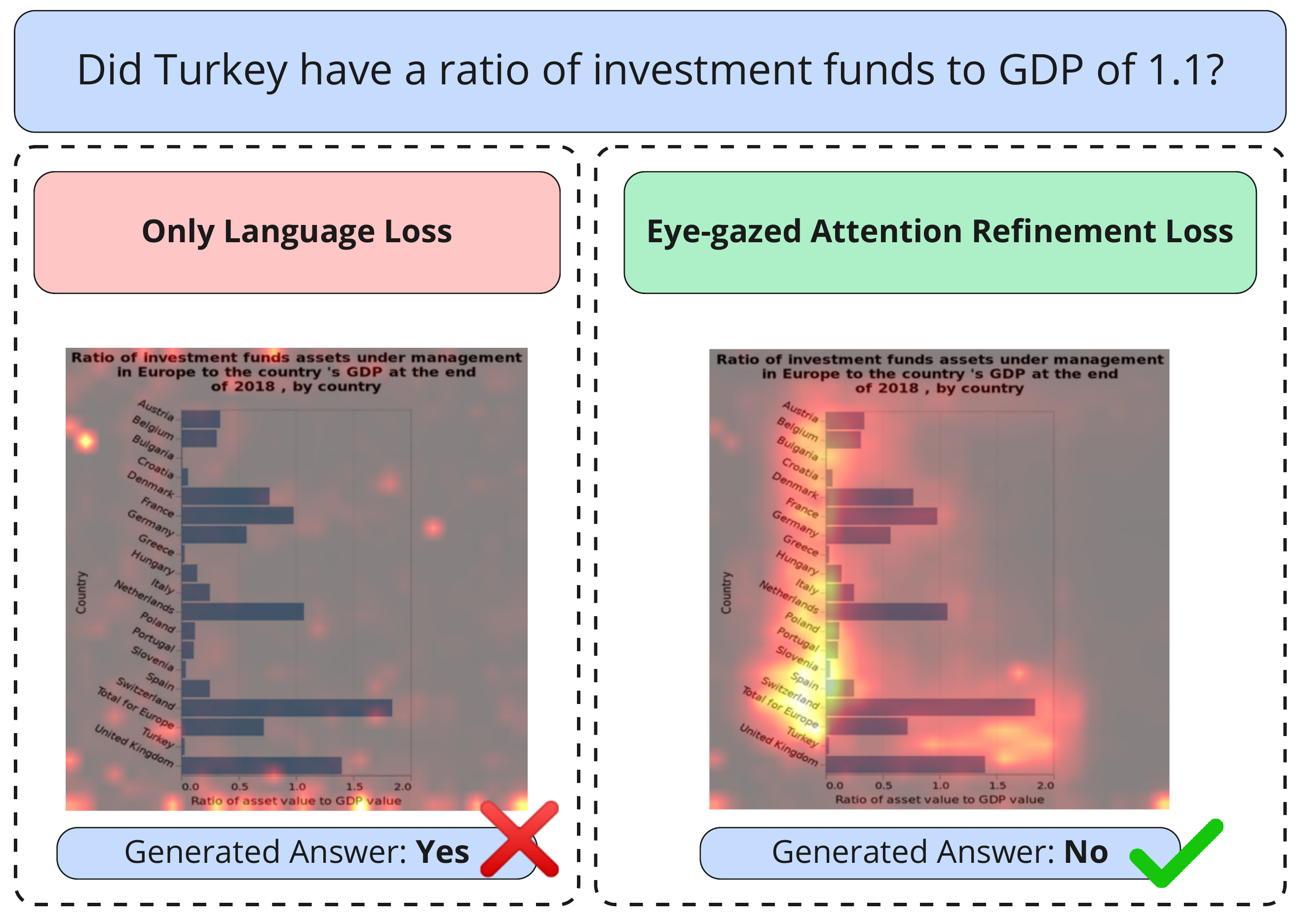}
    \caption{
    {\bf Motivation.} Example of the model's attention maps. The left model, trained with language loss only, attends inconsistently and gives the wrong answer. The right model, trained with our proposed attention refinement loss, focuses more meaningfully on relevant regions and answers correctly.}
    \label{fig:teaser}
\end{figure}

\section{Introduction}
Charts are a common visual medium for communicating structured information and supporting analysis, comparison, and decision-making across domains. With the advancement of large language models (LLMs)~\cite{brown2020language, achiam2023gpt, touvron2023llama, vicuna2023,grattafiori2024llama, anil2023palm,liu2024deepseek} and large vision-language models (LVLMs)~\cite{liu2023visual, team2023gemini, zhou2024tinyllavaframeworksmallscalelarge,chen2024internvl, chen2024far, masry2024chartgemmavisualinstructiontuningchart}, Chart Question Answering (CQA) has emerged as a key research challenge at the intersection of language, vision and data understanding~\cite{masry2022chartqa}. Early CQA approaches often converted charts into structured data or templates, but recent efforts have increasingly focused on leveraging LVLMs directly on chart images~\cite{han2023chartllamamultimodalllmchart, masry2024chartgemmavisualinstructiontuningchart, Tian_2025}.

Automated chart understanding can significantly aid evidence-based decision-making, by supporting fact-checking~\cite{akhtar2023reading} or improving accessibility for visually impaired users through textual or spoken descriptions~\cite{choi2019visualizing}. As these models continue to advance and find increasing deployment, there is an increasing interest in understanding their internal mechanisms, especially their attention dynamics. Attention between visual and textual input modalities plays a vital role in how LVLMs form inductive biases and make an inference. Additionally, image-text attention serves as a window into the model's reasoning process, offering a potential avenue for interpretability and transparency.  This is particularly important in high-stakes domains such as finance, medicine, and scientific research, where the ability to interpret model decisions is critical.

However, recent studies have shown that image-text attention maps in LVLMs often fail to align with the regions humans focus on when answering questions, a phenomenon known as attention misalignment~\cite{lifine, An2024AGLAMO, han2025alignclip, shu2025largevisionlanguagemodelalignment}. This misalignment can cause models to fixate on irrelevant text or visual elements, leading to incorrect predictions and reduced interpretability. In chart understanding, this issue can be even more problematic, as key information, such as a specific bar, point, or legend entry, is often small or densely packed. As a result, the decision-making process of LVLM-based CQA systems becomes untrustworthy, limiting their use  %in domains 
where interpretability is critical.

Hence, in this paper, we analyze the \textit{attention misalignment} in the context of CQA. We observe that LVLMs, even those instruction tuned on charts, frequently attend to irrelevant chart elements, leading to incorrect responses and a reduced interpretability. We draw inspiration from human visual attention, as prior work~\cite{gao2022aligning, yan2024voila} has shown that human gaze tends to align with perceived importance. We leverage eye-tracking data as explicit supervision to guide the attention maps of LVLMs. Specifically, we collect eye-tracking data from human participants responding to chart reasoning questions. Using the human gaze data, we train models for CQA to focus on regions where humans typically fixate, thereby improving both alignment and interpretability. As shown in Figure~\ref{fig:teaser}, models trained with our approach produce more interpretable and human-aligned attention maps, leading to more accurate answers. Empirical results show that aligning LVLM attention with human gaze improves CQA accuracy by up to 2.56 percentage points compared to fine-tuning with language loss alone.

\vspace{0.02in}
\noindent
To summarize, our key contributions are:
\vspace{-0.03in}
\begin{itemize}
\setlength\itemsep{0em}
    \item \textbf{Eye-Tracking Chart Dataset:} We introduce a new eye-tracking dataset for CQA, capturing regions users look at while answering chart-related questions, serving as the ground-truth.

    \item \textbf{Analysis of LVLM Attention on Charts:} To the best of our knowledge, we are the first to conduct a systematic study of  attention patterns of LVLMs on chart understanding and analyze how they compare with human gaze. % for the same task. % during chart reasoning.
    
    \item \textbf{Gaze-Guided Attention Refinement:} We develop a training approach that aligns LVLM attention with human gaze, using a gaze-supervised loss. %to guide cross-attention toward human fixation regions.

\end{itemize}

\section{Related Work}

\noindent
\textbf{Chart Question Answering Datasets:}
Document understanding, particularly scientific chart understanding, has gained significant attention in the machine learning community. As a result, various datasets and benchmarks have been developed to accelerate progress and evaluate models in chart understanding, including summarization~\citep{kantharaj2022chart}, question answering~\citep{masry2022chartqa}, explanation generation~\citep{kantharaj2022opencqa}, and  fact-checking~\citep{akhtar2023reading}. Among these, CQA has become a focal point, driven by the rapid advancements of LVLMs. Early benchmarks such as STL-CQA~\citep{singh2020stl}, LEAF-QA~\citep{chaudhry2020leaf}, FigureQA~\citep{kahou2018figureqa}, and DVQA~\citep{kafle2018dvqa} relied on synthetic charts or templated questions. Later efforts like PlotQA~\citep{methani2020plotqa} and ChartQA~\citep{masry2022chartqa} introduced charts from real-world sources, improving the diversity and realism of the data. Recent benchmarks  % have continued this effort and 
pushed the evaluation into open domains~\citep{masry2025chartqapro, wang2024charxiv, liu2024mmc} and more complex, reasoning-intensive understanding tasks~\citep{fan2024pre, xia2024chartx, xu2023chartbench}. Different from prior work, our benchmark uses eye-tracking annotations alongside chart question pairs to measure how well the LVLMs' attention aligns with that of humans. This human-centric design is crucial for bridging the gap between model performance and interpretability.

\vspace{0.05in}
\noindent
\textbf{Gaze Datasets}:
Several studies have collected human gaze data to understand visual attention on charts. \citet{borkin2016memory} collected a dataset of 393 visualizations along with participants' fixation locations during encoding and recognition of the visualizations.
\citet{POLATSEK201826} analyzed human visual attention during task-solving on 30 charts. More recently, \citet{Shin2022ASD} gathered human attention on 10,960 chart images for the task of chart type recognition using webcam-based eye tracking. In the context of chart question-answering, \citet{wang24_chi} used BubbleView~\citep{Kim2017bubbleview} to crowd-source mouse-click approximations of attention over 3,000 visualizations. While these datasets have contributed valuable insights, many either remain relatively small in size or rely on indirect, lower-fidelity methods such as mouse clicking or webcam-based tracking. We chose to use high-precision eye-tracking equipment for our dataset, as prior work has demonstrated that eye tracking yields more accurate and consistent attention maps than mouse tracking. For example, \citet{tavakoli2017saliencyrevisitedanalysismouse} showed that visual congruency across participants is considerably higher with eye-tracking data, noting that even a large volume of mouse-tracking data from 90 participants could not match the performance of eye-tracking data from just 15 participants. Moreover, eye tracking captures immediate, cognitively grounded attention, whereas mouse-based methods introduce delays due to the slower nature of cursor movement. Prior work has shown that gaze data reliably reflects natural visual behavior and is widely used in saliency prediction, cognitive studies, and attention-aware interface designs  \citep{JacobEyetracking2003, Judd2012ABO, NielsenEyetrackingWeb2009, Majaranta2014}. Therefore, for a deeper analysis of LVLM and human attention and to fine-tune models for improved interpretability, we provide a high-quality, large-scale eye-tracking and CQA dataset (4,638 attention maps).%  in chart understanding tasks. 

\vspace{0.05in}
\noindent
\textbf{Visual Attention in LVLMs:} In LVLMs, the attention mechanism between visual and textual tokens is critical not only for enabling cross-modal interaction but also for providing interpretability into how models integrate visual and textual information~\citep{aflalo2022vl, stan2024lvlm}. Despite this, recent studies have shown that LVLMs often exhibit unintuitive visual attention patterns \citep{arif2025hired, woo2024don}. Specifically, LVLMs tend to assign disproportionately high attention weights to specific tokens which are irrelevant to the text query~\citep{zhang2025mllms, kang2025see}. Moreover, prior works~\citep{liu2024paying, chen2024image, tong2024cambrian} have found that LVLMs under-utilize the visual inputs, leading to text-biased reasoning and weak visual grounding. 

Several techniques have been proposed to address these issues. VAR~\citep{kang2025see} redistributes attention from irrelevant to relevant visual tokens. Visual contrastive decoding~\citep{leng2024mitigating} encourages reliance on visual inputs by contrasting outputs with and without images. Other methods~\citep{zhu2024ibd, zhang2024prompt} explicitly boost attention to visual tokens to improve grounding. Unlike these approaches, we propose a novel approach that guides LVLM attention in training, using human gaze as an implicit supervisory signal, offering a cognitively grounded prior that enhances both reasoning and interpretability.

\begin{figure*}[!ht]
  \centering
  \includegraphics[
    width=\textwidth,
    height=\textheight,
    keepaspectratio
  ]{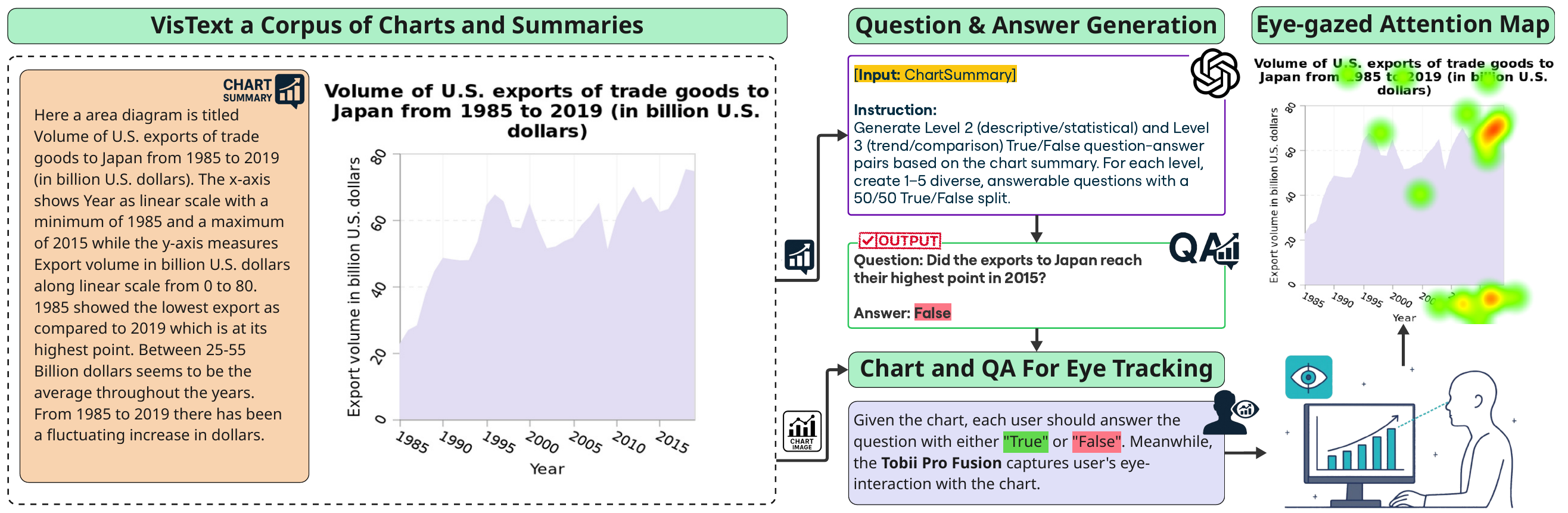}
  \caption{{\bf Overview of the dataset creation pipeline} for generating chart-question-answer pairs with human gaze.}
  \label{fig:dataset}
\end{figure*}

\section{ChartGaze Dataset}
To address attention misalignment in LVLMs \citep{shu2025largevisionlanguagemodelalignment}, we introduce ChartGaze, a novel dataset that captures how humans visually process charts. By capturing high-precision human gaze patterns for chart-based questions, ChartGaze enables detailed comparisons between human and model attention. This opens up new insights into how human gaze can serve as an implicit training signal to enhance LVLM performance on chart understanding. We describe the dataset’s curation process and statistics in the following subsections.

\subsection{Dataset Construction}
ChartGaze builds on chart images from the VisText and ChartQA datasets~\citep{tang2023vistextbenchmarksemanticallyrich, masry2022chartqa}, which feature real-world charts sourced from platforms such as \href{https://www.statista.com/}{Statista} and \href{https://www.pewresearch.org/}{Pew Research}. These charts span a broad range of topics, making them suitable for diverse question generation and visual reasoning.
In what follows, we detail our two-stage dataset construction process: question-answer generation and gaze data collection.

\subsubsection{QA Generation}
While ChartQA already includes queries alongside each chart, for the VisText subset, we generated 3–5 question–answer pairs per chart caption using the pipeline illustrated in Figure~\ref{fig:dataset}. Because VisText summaries are human-authored, detailed, and semantically rich, they provide the necessary context for LLMs to generate questions that are meaningfully grounded in chart content. We used few-shot prompting with GPT-4o to generate questions regarding descriptive statistics, point-wise comparisons, and trend analysis. To ensure high-quality supervision, we instructed the model to:
\vspace{-0.05in}
\begin{itemize}
\setlength\itemsep{0em}
    \item Use diverse phrasing in the questions,
    \item Balance True/False answers equally,
    \item Return output in a strict \texttt{JSON} format to facilitate automatic parsing and validation.
\end{itemize}
The complete prompt can be found in Appendix~\ref{appendix:question_generation}.

For detailed error analysis and to better understand our dataset's composition, we categorized each question into one of six semantic types: Trend Analysis (TA), Finding Extremum (FE), Filtering (F), Comparison (CP), Retrieving Value (RV), and Computing Derived Value (CV). We defined each category with examples and incorporated them into the GPT-4o prompt (Appendix~\ref{appendix:category_prompt}).

\subsubsection{Gaze Collection}
We first detail the UI layout and then the gaze collection process.

Each chart was displayed at the bottom center of the screen, with the corresponding question in the top-left corner. This encouraged participants to read the question first and then shift their gaze to the chart, which helped reduce noise from repeated top-to-bottom transitions. A visualization of our UI is provided in Appendix~\ref{appendix:ui_setup}.

We captured gaze data using the \href{https://www.tobii.com/products/eye-trackers/screen-based/tobii-pro-fusion}{Tobii Fusion Pro} eye-tracker, which records gaze positions at the pixel level with microsecond temporal resolution. 
We then extracted fixation points and aggregated them into gaze maps via the following steps:
\vspace{-0.07in}
\begin{itemize}
\setlength\itemsep{0em}
    \item Compute total fixation duration for each pixel.
    \item Apply a logarithmic non-linearity to smooth out sharp differences in fixation time. 
    \item Apply Gaussian filter on the fixation map to simulate human visual receptive field.
\end{itemize}

\subsection{Quality Control}

We performed a two-step quality control process on our collected data to ensure reliability and accuracy. This included filtering both the generated questions-answers, and the collected gaze data.

\subsubsection{QA Pair Quality Control}
To ensure the correctness of our generated questions, we had human annotators flag ambiguous or incorrect questions during data collection. Of 4,811 questions, only 98 were flagged (a 2.0\% error rate), which is lower than the 3.7\% error rate found in the 593 selected ChartQA questions (22 errors). All flagged instances were removed.

We also verified the quality of the GPT-generated answers by measuring agreement with human annotators. A random sample of 100 examples showed a 93\% agreement between GPT-4o and the majority vote of our seven human annotators. The highest pairwise agreement among human annotators was 92\%. For this subset, Fleiss' Kappa was 0.7098, with an average pairwise agreement of 83.75\%. These high agreement scores confirm the quality of both the GPT-4o answers and the human annotations.

\subsubsection{Gaze Data Quality Control}
We implemented a calibration and filtering process to ensure high-quality gaze data. For each participant, we calibrated the device and manually validated its accuracy, proceeding only when there was 0\% data loss, resulting in an average accuracy of 0.42 degrees (16.8 px).

We also filtered the collected data to remove noise. Specifically, we removed invalid gaze samples and non-fixations caused by blinks or head movements. To ensure reliable gaze maps, we also discarded the bottom 3\% of charts based on total viewing time.

\subsection{Dataset Statistics}

After filtering, our dataset consists of 4,638 attention maps derived from 1,620 unique chart images. These maps were collected from 476 yes/no QA pairs sourced from ChartQA and an additional 4,162 pairs generated from 1,144 VisText captions, averaging 3.6 QA pairs per caption. The dataset was divided into a training set of 3,716 maps (80\%) and a validation set of 922 maps (20\%). The eye-tracking data was gathered from 32 student participants (11 volunteers and 21 paid contributors at \$21/hour). The attention maps are distributed across chart types as follows: 2,470 bar charts, 1,100 line charts, 968 area charts, and 100 pie charts.
Figure \ref{fig:question_distribution} shows the distribution of question categories. The prominence of TA and FE questions underscores the dataset's focus on level 2 and 3 semantic reasoning \citep{lundgard2021accessible}. 
Finally, the average question length was 12.2 words.

\section{Gaze-Guided Refinement Method}
\begin{figure*}[!ht]
  \centering
  \includegraphics[
    width=\textwidth,            % full page width
    height=\textheight,       % up to 80% of text height
    keepaspectratio
  ]{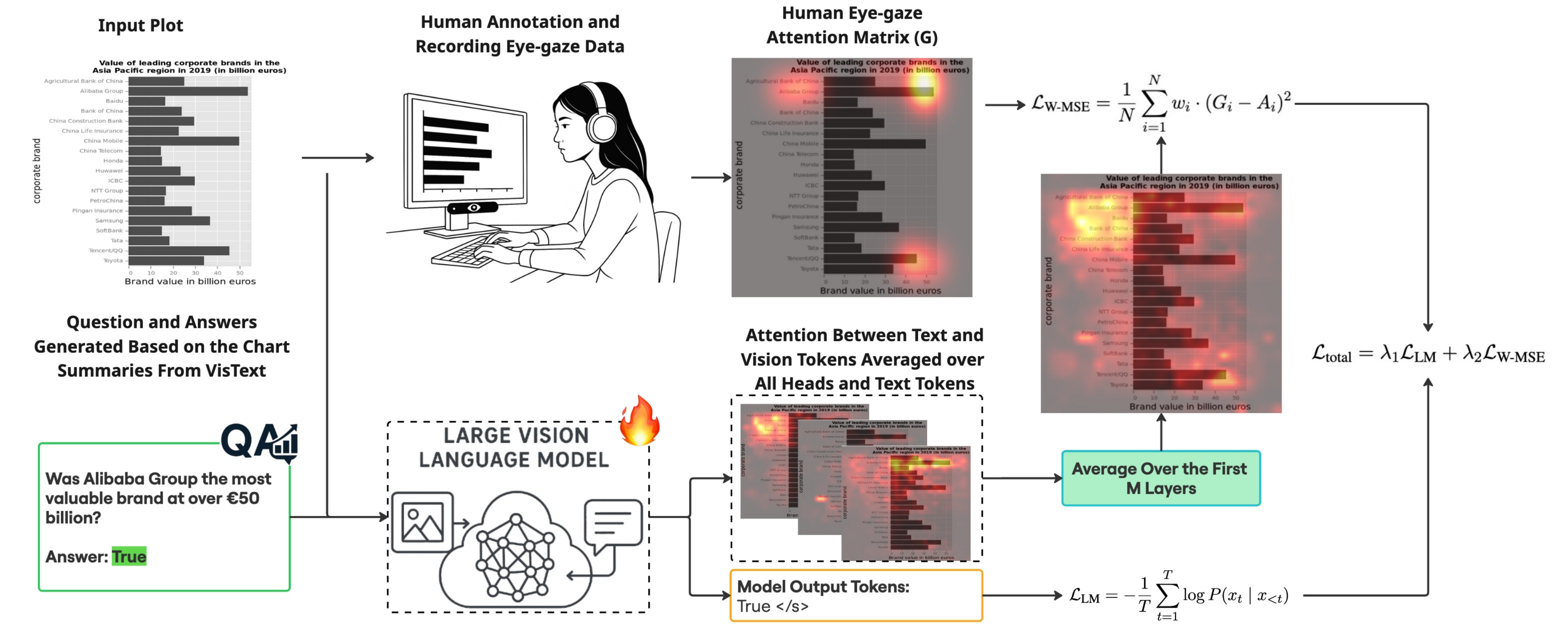}
  \caption{{\bf Overview of our attention refinement training}, including attention map extraction and loss computation.}
  \label{fig:pipeline}
\end{figure*}

LVLMs typically process and integrate information from both visual and textual inputs through multi-head self-attention layers. In this work, we investigate the attention pattern of these models during output generation and tune them to be more interpretable. As shown in Figure~\ref{fig:pipeline}, the attention maps from LVLM are extracted and aligned with human gaze using a joint training objective that combines standard language modelling loss and gaze-guided attention alignment loss. We explain each component in the subsequent sections. 

\subsection{Extracting the Attention Maps}
\label{subsec:extract_attention}
To obtain the image-text attention maps that capture how the model attends to image patches based on the text prompt, we first extract the maps from the first $M$ layers. Our choice to focus on the first $M$ layers is based on a qualitative analysis that aligns with the findings of~\citet{zhang2024redundancy}, which shows that earlier layers of LVLMs are crucial for this type of interaction and that information flow converges in these shallow layers. The relevant part of the attention matrices have a dimension of $\mathbb{R}^{M \times N_h \times T \times I}$, where $M$ is the number of initial layers selected, $N_h$ is the number of heads, $T$ is the number of text tokens, and $I$ is the number of image patches.

To simplify the representation while preserving the core attention structure, we followed the work of
~\citet{jiang2024devils, zhang2024redundancy, helbling2025conceptattention}
and averaged the attention maps across the text tokens, the $N_h$ heads, and the first $M$ layers. We obtained a single aggregated attention score for each image token, represented as $\mathbf{A}' \in \mathbb{R}^{1 \times I}$. This map is then reshaped to the original image dimensions, creating a visual saliency map that is spatially aligned with the gaze data. For more information on our analysis and the specific values of $M$ used for each model, see Appendix~\ref{appendix:implementation_details}.

\subsection{Model Training}
\label{subsec:model_training}

Following the attention extraction process, we train the model to align its visual attention with human gaze patterns while maintaining its language modeling capabilities. To this end, we jointly optimize two objectives: a language modeling loss and a gaze-guided attention alignment loss.

Let $x = \{x_1, \dots, x_{T}\}$ denote the sequence of input tokens representing the question and answer. The standard language modeling loss is defined as:

\begin{equation}
\mathcal{L}_{\text{LM}} = -\frac{1}{T} \sum_{t=1}^{T} \log P(x_t \mid x_{<t})
\end{equation}

To align model attention with human gaze, we used a weighted mean squared error (W-MSE) loss over the flattened attention maps \citep{Bruckert2019DeepSM}. Let $A \in \mathbb{R}^{H \times W}$ denote the model's normalized attention map over the image, and $G \in \mathbb{R}^{H \times W}$ be the corresponding normalized gaze map. We flatten both maps into vectors of length $N = H \times W$: $A = [A_1, A_2, \dots, A_N]$, $G = [G_1, G_2, \dots, G_N]$. For each pixel $i$, we define the weight:

\begin{equation}
w_i = \frac{1}{\alpha - G_i}, \quad \text{with } \alpha = 1.1
\end{equation}

Here, $\alpha$ is a tunable parameter. We followed \citep{Bruckert2019DeepSM} in setting $\alpha$ to $1.1$; this weighting emphasizes areas with higher human gaze values, which we want to prioritize. The W-MSE loss is then:

\begin{equation}
\mathcal{L}_{\text{W-MSE}} = \frac{1}{N} \sum_{i=1}^{N} w_i \cdot (G_i - A_i)^2
\end{equation}

Finally, we combine both objectives (i.e., equations (1) and (3)) into a single loss function:

\begin{equation}
\mathcal{L}_{\text{total}} = \lambda_1 \mathcal{L}_{\text{LM}} + \lambda_2 \mathcal{L}_{\text{W-MSE}}
\end{equation}

Here, $\lambda_1$ and $\lambda_2$ are tunable parameters that we set to 1. This joint training encourages the model to produce gaze-aligned visual attention while preserving its ability to generate accurate and fluent responses.

\section{Experiments and Results}

\begin{figure*}[!ht]
  \centering
  \includegraphics[width=\textwidth]{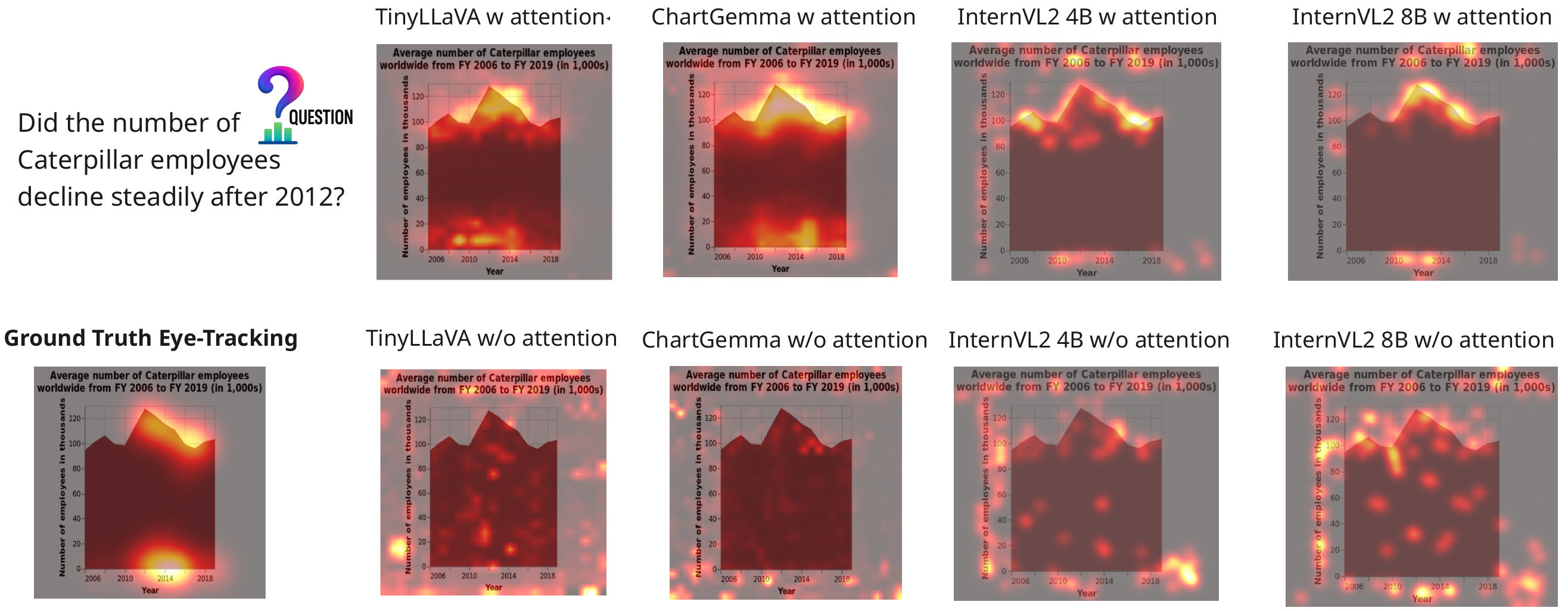}
  \caption{Comparison of attention maps from models trained with attention refinement loss vs. language loss only}
  \label{fig:model_attn_maps}
\end{figure*}

\noindent
\textbf{LVLMs Evaluated:} We evaluated the effectiveness of our approach on four models with diverse architectures, training strategies, and sizes: TinyLLaVA-450M, InternVL2-4B, InternVL2-8B, and ChartGemma-3B. TinyLLaVA-450M is the smallest model in the TinyLLaVA family, which has shown competitive results despite its size \citep{zhou2024tinyllavaframeworksmallscalelarge}. InternVL2 is a state-of-the-art vision-language foundation model with strong performance in visual question answering tasks \citep{chen2024internvl}. We include both the 4B and 8B variants to assess the impact of scale. Finally, ChartGemma-3B is an instruction-tuned model built on PaliGemma and currently represents the state-of-the-art in CQA~\citep{masry2024chartgemmavisualinstructiontuningchart}.

\noindent
\textbf{Metrics:} Since our dataset consists of yes/no questions, we use accuracy, defined as the percentage of questions correctly answered, to compare chart understanding performance across different models. To assess the similarity between the LVLM attention maps and the human attention maps, we used three standard metrics commonly used in saliency prediction~\citep{Meur2012MethodsFC, Bruckert2019DeepSM}. Pearson’s Correlation Coefficient (CC), Kullback–Leibler (KL) divergence, and histogram intersection (also known as similarity or SIM metric) between the ground-truth human gaze distribution and the model's attention distribution over the chart image.

\subsection{Performance Improvement Across Models}

We compare model performance under three conditions: zero-shot, fine-tuning with language loss only, and fine-tuning with our proposed attention-guided loss. As shown in Table~\ref{tab:performance}, models fine-tuned with attention-guided loss consistently outperform those trained solely with language loss. We set the temperature to zero to ensure deterministic zero-shot results. All other experiments were repeated three times, and we report the mean and standard deviation of each evaluation metric.

Notably, TinyLLaVA-450M, InternVL2-4B, and InternVL2-8B achieved statistically significant improvements of 1.19\%, 1.54\%, and 2.56\% respectively. In contrast, ChartGemma-3B showed a marginal improvement of 0.18\%, which was not statistically significant. We hypothesize that this is due to its extensive prior exposure to a corpus of  122,857 charts during instruction tuning \cite{masry2024chartgemmavisualinstructiontuningchart}, which may have resulted in an effective, though less interpretable, attention structure.

In addition to accuracy gains, our approach improves the alignment between model attention and human gaze. As shown in Table~\ref{tab:performance}, models fine-tuned with our method produce higher CC and SIM scores and lower KL divergence, indicating more human-aligned and interpretable attention. Furthermore, we observe a consistent positive correlation between interpretability metrics and QA performance, both within repeated runs of the same model and across different model architectures as shown in Figures~\ref{fig:cc_vs_accuracy_runs} and~\ref{fig:cc_vs_accuracy_models} (in Appendix).

Figure~\ref{fig:model_attn_maps} compares attention maps produced by different models trained with and without our proposed attention refinement loss. Models fine-tuned with language loss alone exhibit noisy attention, often failing to align with salient chart regions. In contrast, models trained with attention supervision display sharper, more human-like focus patterns. ChartGemma, in particular, closely aligns with human fixation maps, while TinyLLaVA similarly produces coherent attention, often emphasizing trend-relevant regions. InternVL2 variants also demonstrate focused activation on key visual elements such as chart peaks and axis labels. These qualitative results support the effectiveness of our approach in guiding models to attend to the most relevant regions of the chart.

\begin{table*}[t]
\centering
\setlength{\tabcolsep}{10pt}
\renewcommand{\arraystretch}{1.2}
\resizebox{\textwidth}{!}{%
\begin{tabular}{llcccc}
\toprule
\textbf{Training} & \textbf{Model} & \textbf{Test Acc.} & \textbf{CC} $\uparrow$ & \textbf{KL} $\downarrow$ & \textbf{SIM} $\uparrow$ \\
\midrule
\multirow{3}{*}{\textbf{Zero-shot}} 
  & TinyLLaVA-450M    &  46.64             & -0.078              &  1.810             & 0.267              \\
  & InternVL2-4B       & 49.86        & -0.060           & 1.722         & 0.282 \\
    & InternVL2-8B       & 50.93        & -0.054           & 1.681         & 0.296 \\
  & ChartGemma-3B   & \underline{52.39}              &  \underline{0.100}             & \underline{1.559}              & \underline{0.323}              \\
\midrule
\multirow{3}{*}{\textbf{Without Attn}} 
  & TinyLLaVA-450M    & 62.58 $\pm$ 0.27            & -0.048 $\pm$ 0.005             & 1.705 $\pm$ 0.031              & 0.288  $\pm$ 0.004           \\
  & InternVL2-4B       & 63.91 ± 0.20 & -0.028 ± 0.004   & 1.532 ± 0.010   & 0.301 ± 0.004 \\
  & InternVL2-8B     &      65.36 $\pm$ 0.22        &      -0.017 $\pm$ 0.003         &        \underline{1.487 $\pm$ 0.009}       &      	0.312 $\pm$ 0.004         \\
  & ChartGemma-3B   &     \underline{72.49 $\pm$ 1.69}         &      \underline{0.092 $\pm$ 0.004}       &   1.594 $\pm$ 0.026           &    \underline{0.316 $\pm$ 0.003}          \\
\midrule
\multirow{3}{*}{\textbf{With Attn}} 
  & TinyLLaVA-450M    & 63.77 $\pm$ 0.54             & 0.391 $\pm$ 0.007              & 1.132 $\pm$ 0.015 & 0.439 $\pm$ 0.002           \\
  & InternVL2-4B       & 65.45 ± 0.23 & 0.402 ± 0.006   & 1.072 ± 0.008   & 0.451 ± 0.004 \\
  & InternVL2-8B     &       67.92 $\pm$ 0.15        &        0.417 $\pm$ 0.006       &       1.036 $\pm$ 0.007        &       \textbf{0.468 $\pm$ 0.005 }       \\
  & ChartGemma-3B   &   \textbf{72.67 $\pm$ 1.24}            &  \textbf{ 0.436 $\pm$ 0.011}            &  \textbf{1.033 $\pm$ 0.014 }          &  0.452 $\pm$ 0.005         \\
\bottomrule

\end{tabular}
}
\caption{Performance of models trained with and without attention loss. $\uparrow$ / $\downarrow$ indicates higher / lower is better.}
\label{tab:performance}
\end{table*}

\newcommand{\halfvrule}{\rule[0.5ex]{0.4pt}{1.5ex}}

\begin{table}[t]
\centering
\setlength{\tabcolsep}{4pt}
\resizebox{\linewidth}{!}{%
\begin{tabular}{lcccc}
    \toprule
    \textbf{Condition} & \textbf{Acc.} $\uparrow$ & \textbf{CC} $\uparrow$ & \textbf{KL} $\downarrow$ & \textbf{SIM} $\uparrow$ \\
    \midrule
    \textbf{Language loss only:} & 65.36 & -0.017 & 1.487 & 0.312 \\
    \quad\vrule--- Blur human gaze areas & 61.02 & -0.139 & 1.681 & 0.236 \\
    \quad\vrule--- Mask human gaze areas & 60.14 & -0.124 & 1.713 & 0.221 \\
    \quad\vrule--- Blur non-gaze areas & 64.10 & 0.112 & 1.392 & 0.298 \\
    \quad\halfvrule--- Mask non-gaze areas & 62.85 & 0.064 & 1.456 & 0.274 \\
    \addlinespace
    \midrule
    \textbf{Gaze supervision + language loss:} & \textbf{67.92} & \textbf{0.417} & \textbf{1.036} & \textbf{0.468} \\
    \quad\vrule--- Blur human gaze areas & 60.84 & -0.174 & 1.794 & 0.201 \\
    \quad\vrule--- Mask human gaze areas & 59.92 & -0.152 & 1.752 & 0.188 \\
    \quad\vrule--- Blur non-gaze areas & 66.82 & 0.284 & 1.218 & 0.395 \\
    \quad\halfvrule--- Mask non-gaze areas & 63.72 & 0.203 & 1.314 & 0.356 \\
    \bottomrule
    \end{tabular}
}
\caption{Ablations showing the importance human-aligned attention towards the model's reasoning process. The experiments are grouped by training setup. Each header row reports the unperturbed model’s performance; indented rows apply perturbations to probe model's reliance on attended vs.\ non-attended regions.}
\label{tab:mask_performance}
\end{table}

\subsection{Error Analysis}
\begin{figure}[t]
    \centering
    \includegraphics[width=\linewidth]{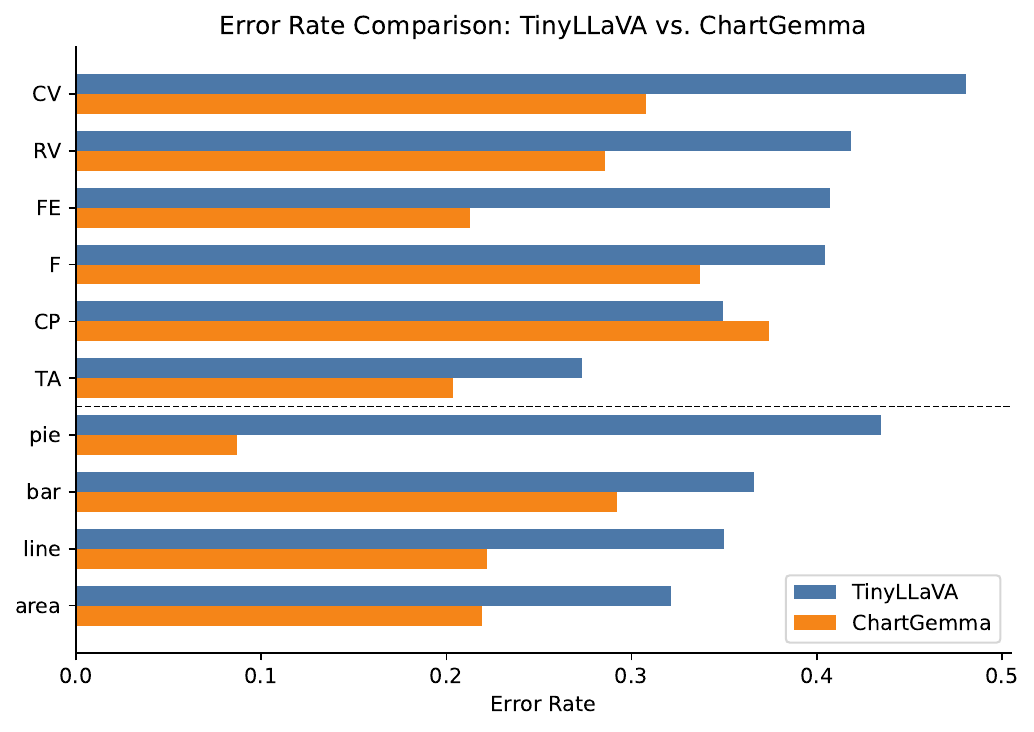}
    \caption{
    Comparison of error rates for TinyLLaVA and ChartGemma across question categories (top) and chart types (bottom), sorted by TinyLLaVA's error rates. A dashed line separates the two groups. ChartGemma consistently shows lower error rates, particularly on chart types and reasoning-heavy categories.}
    \label{fig:error_comparison}
\end{figure}

We analyze error rates by question categories and chart types for both TinyLLaVA and ChartGemma. As shown in Figure~\ref{fig:error_comparison}, TinyLLaVA struggles the most with computing derived value (CV) questions, which require multi-step reasoning, while it performs best on trend analysis (TA), likely due to the high number of TA examples in the training set.

Across chart types, TinyLLaVA performs relatively consistently, with the exception of pie charts, where it shows notably higher error. This may be attributed to limited exposure; only 77 pie chart examples were included in training.

ChartGemma outperforms TinyLLaVA in most categories, except for TA and comparison (CP) questions. This may be because %of 
its extensive prior exposure to over 120K charts during instruction tuning limited its responsiveness to the specific supervision in our setting.

\subsection{Ablation Studies: Masked Inference}
To evaluate the role of human-aligned attention in the model’s reasoning process, we conducted controlled ablation experiments using two intervention strategies: masking and blurring human-attended regions. We used the fine-tuned InternVL2-8B model on our human gaze data in this experiment.

For the masking condition, we generated a binary mask from human gaze annotations, where pixels attended by humans were set to 0. For the blurring condition, we directly modified the input images by applying a Gaussian blur (kernel size = 15, $\sigma$ = 5) to the regions with high human attention density, thus degrading the visual information in those critical areas. These experiments allow us to assess whether the model merely mimics human attention or leverages it for accurate reasoning.

As shown in Table \ref{tab:mask_performance}, performance dropped significantly for the attention-tuned models, with a 7.08\% decrease for blurring and an 8.00\% decrease for masking. In contrast, the model trained only with a language loss saw a smaller drop of 4.34\% and 5.22\%, respectively. This suggests that the attention-tuned model relies more heavily on these semantically meaningful areas to generate its answers. Blurring the non-human-attended regions, on the other hand, resulted in a relatively minor performance decrease (1.90\%), reinforcing that the model is not only aligned with human gaze visually but also actively uses that information during reasoning.

\subsection{Ablation Studies: Loss Function}
We investigated the role of different loss functions in aligning TinyLLaVA-450M attention maps with human gaze data. Specifically, we chose our losses from two broad categories of loss functions: pixel-based and distribution-based.

For pixel-based loss, we adopted W-MSE as defined in Section \ref{subsec:model_training}, which prioritizes regions with high fixation density. Among distribution-based losses, we evaluated KL Divergence (KLD), Focal Loss \citep{lin2018focallossdenseobject}, and a combined Dice + Binary Cross Entropy (BCE) loss. KLD encourages similarity between the extracted and ground-truth human attention distributions, while Focal Loss and Dice + BCE emphasize salient regions with higher ground-truth attention values. Detailed information is provided in Appendix~\ref{appendix:loss}.

Figure~\ref{fig:loss_ablation} in Appendix presents qualitative comparisons of attention maps extracted from models trained with each loss function. As can be seen, W-MSE matches the ground truth most closely. KL and Focal loss have resulted in uniformly high attention on the axis and top of the bar and BCE + Dice had a too focused attention map that misses some critical points. 

Table~\ref{tab:loss_comparison} reports test accuracy along with three attention evaluation metrics. W-MSE yields the best performance, achieving both the highest task accuracy and the most faithful attention alignment with human gaze. Notably, we observe a consistent trend across loss types: improvements in attention quality are correlated with gains in task accuracy.

\begin{table*}[th]
\centering
\small
\begin{tabular}{lcccc}
    \toprule
    \textbf{Training Setup} & \textbf{Test Acc.} & \textbf{CC $\uparrow$} & \textbf{KL $\downarrow$} & \textbf{SIM $\uparrow$} \\
    \midrule
    \textbf{Without Attn Supervision} \\
    \quad \textsc{model-25\%} & 60.21 $\pm$ 0.73 & -0.045 $\pm$ 0.005 & 1.602 $\pm$ 0.012 & 0.274 $\pm$ 0.006 \\
    \quad \textsc{model-50\%} & 63.58 $\pm$ 0.24 & -0.028 $\pm$ 0.004 & 1.530 $\pm$ 0.010 & 0.295 $\pm$ 0.005 \\
    \quad \textsc{model-100\%} & \underline{65.36 $\pm$ 0.22} & \underline{-0.017 $\pm$ 0.003} & \underline{1.487 $\pm$ 0.009} & \underline{0.312 $\pm$ 0.004} \\
    \midrule
    \textbf{With Attn Supervision} \\
    \quad \textsc{model-25\%} & 64.07 $\pm$ 0.26 & 0.297 $\pm$ 0.008 & 1.174 $\pm$ 0.011 & 0.402 $\pm$ 0.006 \\
    \quad \textsc{model-50\%} & 66.51 $\pm$ 0.20 & 0.396 $\pm$ 0.007 & 1.065 $\pm$ 0.009 & 0.454 $\pm$ 0.005 \\
    \quad \textsc{model-100\%} & \textbf{67.92 $\pm$ 0.15} & \textbf{0.417 $\pm$ 0.006} & \textbf{1.036 $\pm$ 0.007} & \textbf{0.468 $\pm$ 0.005} \\
    \bottomrule
\end{tabular}
\caption{Performance of models trained with and without attention loss across different dataset sizes.}
\label{tab:attn_supervision_results}
\end{table*}

\begin{table}[t]
\centering
\resizebox{\linewidth}{!}{%
\begin{tabular}{lcccc}
\toprule
\textbf{Loss Function} & \textbf{Test Acc.} & \textbf{CC} $\uparrow$ & \textbf{KL} $\downarrow$ & \textbf{SIM} $\uparrow$ \\
\midrule
W-MSE       & \textbf{64.53} & \textbf{0.386} & \textbf{1.145} & \textbf{0.438} \\
KLD         & 62.36 & 0.306 & 1.209 & 0.380 \\
Focal Loss  & 61.06 & 0.339 & 1.188 & 0.388 \\
Dice + BCE  & 60.41 & 0.194 & 4.174 & 0.183 \\
\bottomrule
\end{tabular}
}
\caption{Impact of loss functions on performance. }
\label{tab:loss_comparison}
\end{table}

\begin{table}[th]
\centering
\small
\begin{tabular}{ccccc}
\toprule
\textbf{Fixation $\sigma$} & \textbf{Test Acc.} & \textbf{CC $\uparrow$} & \textbf{KL $\downarrow$} & \textbf{SIM $\uparrow$} \\
\midrule
\quad \textsc{20} & 62.89 &  0.277& 1.875 & 0.272\\
\quad \textsc{40} & \textbf{63.77}  & 0.391 & 1.132 & 0.439  \\
\quad \textsc{80} & 61.49 & \textbf{0.490} & \textbf{0.588} & \textbf{0.612}\\

\bottomrule
\end{tabular}
\caption{Performance of models trained with attention loss across different $\sigma$ values.}
\label{tab:sigma_analysis}
\end{table}

\subsection{Effect of Dataset Size}
In domains where interpretability is essential, such as medical applications, training data is often limited. To evaluate the effectiveness of our attention supervision approach under low-data settings, we conducted experiments using randomly selected 25\% and 50\% of the ChartGaze dataset in addition to our previous result on the entire dataset. 
We trained InternVL2-8B with three different random seeds for each data setting to account for variability, which tends to be amplified in low-data regimes. This is reflected in the higher standard deviations reported in Table~\ref{tab:attn_supervision_results}.
Our method consistently outperforms models trained with standard language supervision, with performance gains becoming more pronounced as data availability decreases. Specifically, our approach achieves accuracy improvements of 3.86, 2.93, and 2.56 percentage points for the 25\%, 50\%, and 100\% subsets, respectively. This trend highlights the value of attention supervision in low-resource settings.

Moreover, we again observe a strong linear correlation between attention quality and task accuracy, reinforcing the link between interpretable attention mechanisms and overall model performance.

\subsection{Gaze Map Post-Processing Analysis}
As part of our gaze map post-processing pipeline, we applied a Gaussian filter with a fixed standard deviation of $\sigma = 40$ pixels, corresponding to the spatial spread of visual attention around fixation points. To investigate the sensitivity of our model to this hyperparameter, we trained TinyLLaVA-450M with two different values of $\sigma$. Table \ref{tab:sigma_analysis} presents the results of this analysis. Our choice of sigma yields the best accuracy. Interestingly, $\sigma = 80$ scores really well on the attention quality metrics, but has a lower accuracy. This is because choosing such a high sigma results in high attention in a very large radius, therefore the model learns to have high attention in many regions, which is not beneficial for distinguishing  the relevant parts of the image. Thus, high attention on a large part of the image as shown in Figure \ref{fig:sigma-choice}, results in a good metric value but in this case it is not same as learning which regions to attend to and hence not improving the accuracy.

\section{Conclusion and Future Work}

As chart understanding models are increasingly deployed in real-world applications, it is critical to ensure their interpretability. In this work, we introduced a novel attention refinement method and demonstrated its effectiveness on the newly collected ChartGaze dataset. Our results show that attention supervision not only improves alignment with human gaze (making the model attend to the interpretable parts of the chart) but also leads to performance gains in non-instruction-tuned models. While our method proved effective across multiple architectures, including TinyLLaVA and InternVL, it yielded only marginal gains on ChartGemma. This may be due to ChartGemma’s prior instruction tuning on chart-related tasks, which could result in attention patterns that are either already well-formed or less responsive to additional supervision. Future work could explore strategies to better integrate attention refinement into instruction-tuned models, as well as extend this work to more diverse chart types and free-form question formats to better understand how attention varies with task complexity.

\noindent
\section*{Limitations}
This study focuses on Yes/No questions and relatively simple chart types (bar, line, and pie). These choices enabled large-scale data collection while allowing for accurate capture of participant attention during reasoning. However, these decisions may limit the generalizability of our findings to more complex visualizations and open-ended questions, where attention behavior may differ.

\section*{Ethical Considerations}
This study was reviewed and approved by the University of British Columbia behavioural research ethics board.
All participants provided informed consent prior to participation. Participants were recruited based on normal vision and hearing criteria and completed a chart interpretation task while eye-tracking data were collected. To protect privacy, all collected data were de-identified and securely stored on encrypted devices. Only anonymized data were used for analysis and model training. Participants were compensated at a rate of \$7 per 20-minute session, up to a maximum of \$21. They were informed of their right to withdraw at any time without penalty. Any public release of the dataset ensures participant anonymity and complies with open-access research guidelines. Moreover, our use of ChartQA and VisText is consistent with their intended purpose as open research datasets for chart understanding. For the ChartGaze dataset we created, we specify its intended use is for research on chart understanding, question-answering and interpretability. Model trained using our approach may still have over-reliance on superficial visual-textual correlations, leading to plausible-sounding but incorrect answers. We caution against deploying this system in high-stakes environments without robust safeguards and proper fact checking. We used generative AI tools, including ChatGPT, to support editing, formatting, and idea refinement during the research and writing process. All intellectual contributions, experimental designs, and analyses were developed and validated by the authors. No AI-generated content was included without human review and revision.

\section*{Acknowledgments}
We gratefully acknowledge the support of Google for providing computing credits used in this work.
This work was funded, in part, by the Vector Institute for AI, Canada CIFAR AI Chairs, NSERC Canada Research Chair (CRC), and NSERC Discovery. Resources used in preparing this research were provided, in part, by the Digital Research Alliance of Canada and by John R. Evans Leaders Fund CFI grant.

% Bibliography entries for the entire Anthology, followed by custom entries
%\bibliography{anthology,custom}
% Custom bibliography entries only
\bibliography{custom}

\clearpage
\appendix
\onecolumn
\section{Prompt for Question-Answer Generation}
\label{appendix:question_generation}

\begin{tcolorbox}[breakable, title=Question and Answer Generation Prompt]

Generate a set of question-answer pairs based on the following summary of the chart in JSON format.
Focus on the following semantic levels of questions:

\begin{itemize}
    \item \textbf{Level 2 (L2):} Descriptive statistics, extrema, outliers, and correlations (binary: True/False)
    \item \textbf{Level 3 (L3):} Point-wise comparisons, complex trends, pattern synthesis (binary: True/False)
\end{itemize}

For \textbf{Level 2} and \textbf{Level 3}, make decisions on how to generate the questions to produce specific True/False answers. Ensure that the generated questions:
\begin{itemize}
    \item Have a \textbf{50/50 chance} of being True or False.
    \item For False answers, use values that are \textbf{domain-appropriate and plausible} based on the chart data.
\end{itemize}

Use diverse language and ensure that the questions are relevant to the key points in the text and that the answers are accurate and concise. Only include questions that have a clear answer in the text. Aim to generate 1--5 questions for each level.

\textbf{IMPORTANT:} Return your response as a valid JSON array where each question follows this exact format:
\begin{verbatim}
[
  {
    "question": "...?",
    "answer": true/false,
    "level": 2/3
  },
  ...more questions...
]
\end{verbatim}

\textbf{VERY IMPORTANT:} DO NOT use markdown code blocks (```json or ```). Return ONLY the raw JSON.

There are five chart summary, QA pairs below as examples. Note that level 1 is given for the context of the chart and we don't want to create question answer pairs for this level.

---

\textbf{First example:} \\
\emph{Summary:} \\
L1: This line diagram is titled Canadian imports of bauxite from 2005 to 2019 (in 1,000 metric tons). The x-axis measures Year on linear scale from 2006 to 2018 while the y-axis shows Imports in thousand metric tons on linear scale of range 0 to 4,000. \\
L2L3: The year with the lowest import of bauxite was 2009. Although the results fluctuate from year to year, the graph tends to show a general increase in bauxite imports over time, with the exception of 2009, where there was a large decrease in imports.

\emph{QAs (in JSON format):}
\begin{verbatim}
[
  {
    "question": "Was the year with the lowest import of bauxite 2009?",
    "answer": true,
    "level": 2
  },
  {
    "question": "Was the year with the highest import of bauxite 2008?",
    "answer": false,
    "level": 2
  },
  {
    "question": "Did the Canadian imports of bauxite tend to show a general increase 
    over time?",
    "answer": true,
    "level": 3
  },
  {
    "question": "Was there a large decrease in imports in 2015?",
    "answer": false,
    "level": 3
  }
]
\end{verbatim}

---

\textbf{Second example:} \\
\emph{Summary:} \\
L1: Poverty rate for families in the United States from 1990 to 2019 is a line chart. The y-axis plots Poverty rate while the x-axis plots Year. \\
L2L3: Poverty in the USA was at its highest in the early nineties, dropping in 2000 to its second lowest point on the chart. Then poverty rose gradually again, almost hitting the same peak in 2010 and staying level for a few years before dropping sharply in the mid 2000's to well below the levels of the 90s.

\emph{QAs (in JSON format):}
\begin{verbatim}
[
  {
    "question": "Was poverty at its highest in the early nineties?",
    "answer": true,
    "level": 2
  },
  {
    "question": "Was the second lowest poverty rate recorded in 2005?",
    "answer": false,
    "level": 2
  },
  {
    "question": "Did the poverty rate rise from 2000 to 2010?",
    "answer": true,
    "level": 3
  },
  {
    "question": "Was there a rise in poverty rates right after 2010?",
    "answer": false,
    "level": 3
  },
  {
    "question": "Was the poverty level in 90s way higher than that of mid 2000s?",
    "answer": true,
    "level": 3
  },
  {
    "question": "Did the poverty in the USA fluctuate over the years but had an 
    overall upward trend?",
    "answer": false,
    "level": 3
  }
]
\end{verbatim}

---

\textbf{Third example:} \\
\emph{Summary:} \\
L1: This bar diagram is labeled Top 10 U.S. states based on production value of principal fresh and processing market vegetables in 2019 (in 1,000 U.S. dollars). There is a categorical scale with Arizona* on one end and Washington at the other on the x-axis, marked State. Production value in thousand U.S. dollars is plotted on the y-axis. \\
L2L3: California has the most production value of principal fresh and processing market vegetables at nearly 8,000,000 US dollars whereas New Jersey has the least production value of principal fresh and processing market vegetables at only approximately 100,000 US dollars.

\emph{QAs (in JSON format):}
\begin{verbatim}
[
  {
    "question": "Was California the state with the highest production value?",
    "answer": true,
    "level": 2
  },
  {
    "question": "Was the production value of New Jersey 500,000 US dollars?",
    "answer": false,
    "level": 2
  },
  {
    "question": "Did New Jersey have the second-highest production value?",
    "answer": false,
    "level": 3
  },
  {
    "question": "Was the difference between the highest and lowest production values
    around 7,900,000 US dollars?",
    "answer": true,
    "level": 3
  }
]
\end{verbatim}

---

\textbf{Fourth example:} \\
\emph{Summary:} \\
L1: This bar diagram is titled Egypt: National debt from 2015 to 2025 in relation to gross domestic product (GDP). The y-axis shows Year on a categorical scale with 2015 on one end and 2025* at the other. Along the x-axis, National debt in relation to GDP is measured with a linear scale of range 0.0 to 1.0. \\
L2L3: The National debt in Egypt between 2015 to 2025 has been pretty consistent with a slight rise in 2017.

\emph{QAs (in JSON format):}
\begin{verbatim}
[
  {
    "question": "Was there a slight rise in national debt in 2017?",
    "answer": true,
    "level": 2
  },
  {
    "question": "Was national debt at its highest in 2017?",
    "answer": false,
    "level": 2
  },
  {
    "question": "Has the national debt in Egypt been consistent overall?",
    "answer": true,
    "level": 3
  },
  {
    "question": "Was there a sharp drop in national debt in 2018?",
    "answer": false,
    "level": 3
  }
]
\end{verbatim}

---

\textbf{Fifth example:} \\
\emph{Summary:} \\
L1: This is an area plot called Number of passengers arriving and departing at airport terminals in the United Kingdom (UK) from 1992 to 2019 (in millions). On the x-axis, Year is measured. Passengers in millions is plotted on the y-axis. \\
L2L3: The number of passengers at UK airports has risen steadily since 1992 to 2019 with the exception of a period of a few years in 2007--2010 where numbers fell. Overall, numbers have nearly tripled from just over 100 million passengers in 1992 to almost 300 million in 2019. The rate of increase has been relatively steady with the exception of the 2007--2010 period.

\emph{QAs (in JSON format):}
\begin{verbatim}
[
  {
    "question": "Was the number of passengers in 1992 just over 100 million?",
    "answer": true,
    "level": 2
  },
  {
    "question": "Did passenger numbers fall sharply from 2015 to 2019?",
    "answer": false,
    "level": 2
  },
  {
    "question": "Did the number of passengers double from 1992 to 2019?",
    "answer": false,
    "level": 3
  },
  {
    "question": "Was the period of 2007-2010 an outlier with falling numbers?",
    "answer": true,
    "level": 3
  }
]
\end{verbatim}

---

Now generate question-answer pairs from the following summary in the specified JSON format:  
\{caption\}
\end{tcolorbox}

\section{Prompt for Question Category Annotation}
\label{appendix:category_prompt}

\begin{tcolorbox}[breakable, title=Question Category Prompt]
Classify the following questions into one of the six categories based on its intent:
\begin{itemize}
    \item CP (Comparison): The question compares two or more data points (e.g., "Which country has more X than Y?").
    \item CV (Computing Derived Value): The question involves computing a value from others (e.g., difference, average, ratio).
- RV (Retrieving Value): The question asks for the value(s) of specific data points or attributes.
    \item FE (Finding Extremum): The question asks for the maximum or minimum value in the data.
    \item F (Filtering): The question asks for data points that satisfy multiple specified conditions.
    \item TA (Trend Analysis): The question asks about changes over time, patterns, increases, decreases, or stability.
\end{itemize}
\textbf{Output format:}  
Return only the category abbreviation without additional text in a list.

\textbf{Questions:}  
\{questions\}
\end{tcolorbox}

\twocolumn
\begin{figure}[h]
    \centering
    \includegraphics[width=0.8\linewidth]{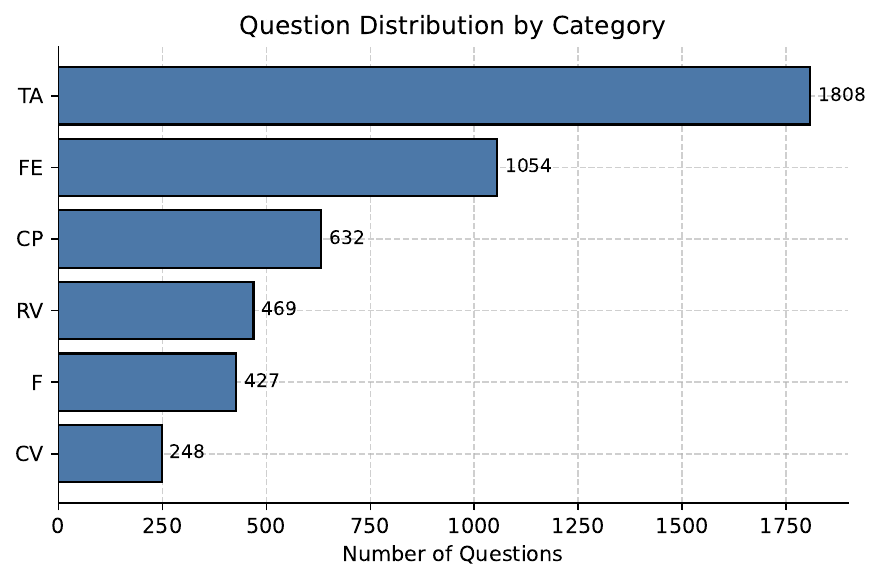}
    \caption{Distribution of questions across categories used in the prompt. TA and FE are the most common, reflecting the reasoning-heavy nature of the dataset.}
    \label{fig:question_distribution}
\end{figure}

\section{User Interface Setup and Participant Demographics}

\begin{figure}[h]
\centering
\includegraphics[width=\linewidth]{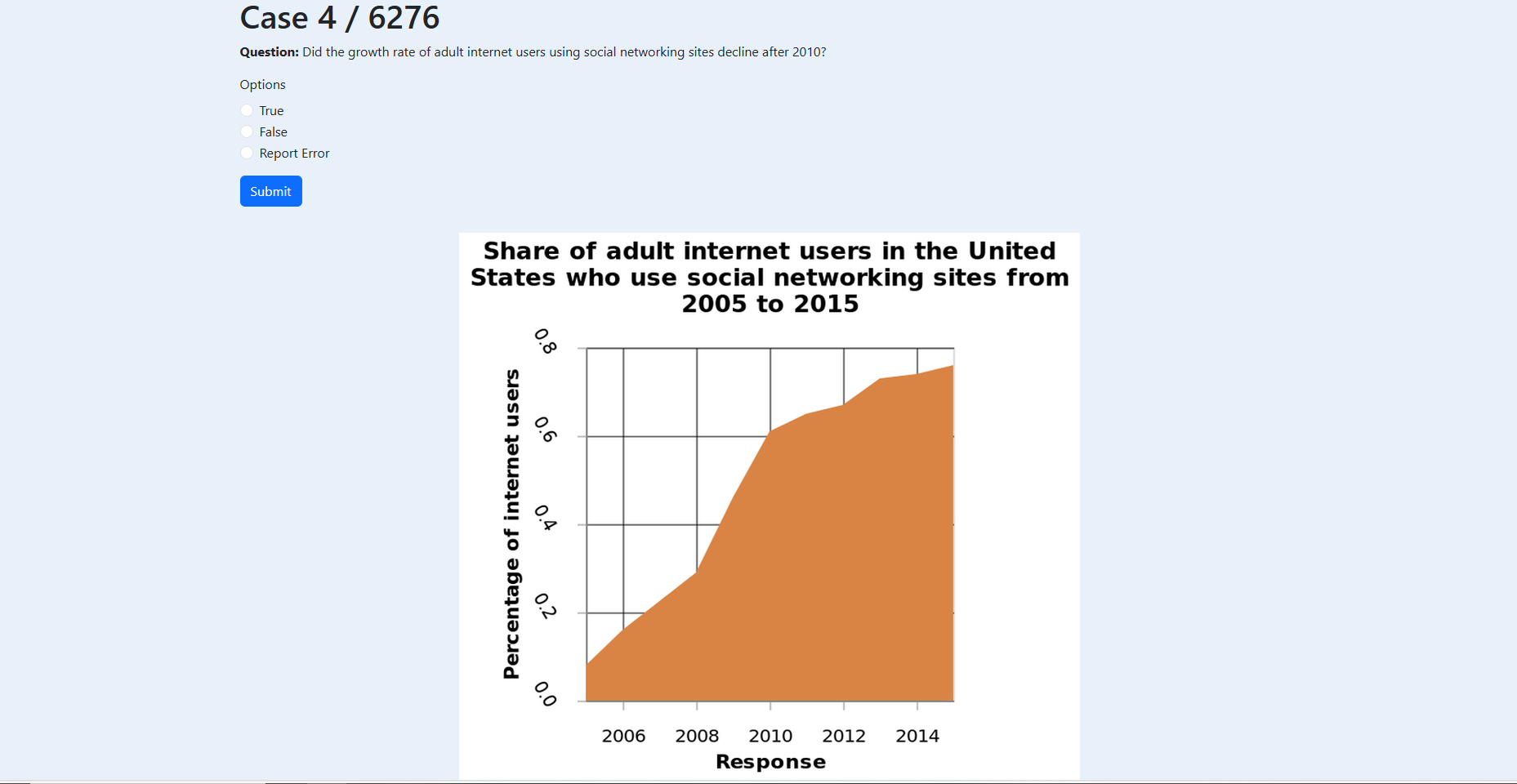}
\caption{Example user interface setup used for collecting gaze data.}
\label{fig:ui_setup}
\end{figure}

Figure~\ref{fig:ui_setup} shows the interface used for our chart-based Yes/No question answering task. All participants were students who reported interacting with charts on a daily to weekly basis. The cohort was gender-balanced and represented a range of academic backgrounds, including engineering, computer science, and statistics.

\label{appendix:ui_setup}

\begin{figure*}[!ht]
  \centering
  \includegraphics[width=\textwidth]{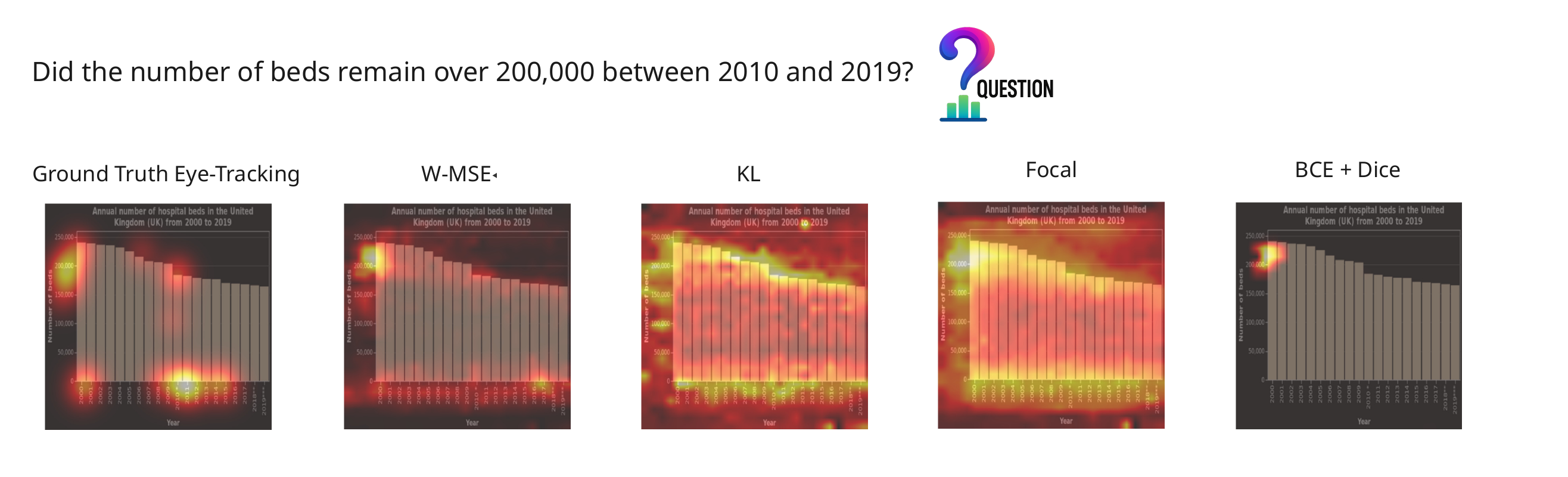}
  \caption{Comparison of attention maps from models trained with different attention refinement loss functions.}
  \label{fig:loss_ablation}
\end{figure*}

\section{Attention Loss Function Details}
\label{appendix:loss}

\paragraph{Focal Loss.}
To address the foreground-background imbalance in gaze prediction, we adopt the Focal Loss introduced by \citet{lin2018focallossdenseobject}. This loss down-weights well-classified examples and focuses training on hard examples. It is defined as:
\begin{multline}
\mathcal{L}_{\text{FL}}(P, Q) = -\sum_{i=1}^{H \times W} \Big[P_i (1 - Q_i)^\gamma \log Q_i \\
+ (1 - P_i) Q_i^\gamma \log (1 - Q_i) \Big]
\end{multline}

\noindent where $i$ indexes $H\times W$ pixels, $P_i$ is the ground truth, $Q_i$ is the predicted value, and $\gamma$ is a focusing parameter (set at 2 in our experiments).

\paragraph{Dice + BCE Loss.}
To improve overlap between predicted and true gaze maps, we combine Dice Loss with Binary Cross-Entropy (BCE). The total loss is:
% \begin{equation}
% \mathcal{L}_{\text{BCE+Dice}}(P, Q) = \lambda_{\text{Dice}} \cdot \mathcal{L}_{\text{Dice}}(P, Q) + \lambda_{\text{BCE}} \cdot \mathcal{L}_{\text{BCE}}(P, Q)
% \end{equation}
% where $\lambda_{\text{Dice}}$ and $\lambda_{\text{BCE}}$ are scalar weights. We use $\lambda_{\text{Dice}} = 100$ and $\lambda_{\text{BCE}} = 1.0$. The Dice loss is defined as:
% \begin{equation}
% \mathcal{L}_{\text{Dice}}(P, Q) = 1 - \frac{2 \sum_i P_i Q_i + \varepsilon}{\sum_i P_i + \sum_i Q_i + \varepsilon}
% \end{equation}
% where $\varepsilon = 10^{-8}$ is added for numerical stability. The BCE loss is given by:
% \begin{equation}
% \mathcal{L}_{\text{BCE}}(P, Q) = - \frac{1}{N \times M} \sum_{i=1}^{N \times M} \left[ P_i \log Q_i + (1 - P_i) \log (1 - Q_i) \right]
% \end{equation}
\begin{multline}
\mathcal{L}_{\text{BCE+Dice}}(P, Q) = \lambda_{\text{Dice}} \cdot \mathcal{L}_{\text{Dice}}(P, Q) \\
+ \lambda_{\text{BCE}} \cdot \mathcal{L}_{\text{BCE}}(P, Q)
\end{multline}

\noindent where $\lambda_{\text{Dice}}$ and $\lambda_{\text{BCE}}$ are scalar weights. We use $\lambda_{\text{Dice}} = 100$ and $\lambda_{\text{BCE}} = 1.0$. The Dice loss is defined as:

\begin{equation}
\mathcal{L}_{\text{Dice}}(P, Q) = 1 - \frac{2 \sum_i P_i Q_i + \varepsilon}{\sum_i P_i + \sum_i Q_i + \varepsilon}
\end{equation}

\noindent where $\varepsilon = 10^{-8}$ is added for numerical stability. The BCE loss is given by:

\begin{multline}
\mathcal{L}_{\text{BCE}}(P, Q) = - \sum_{i=1}^{H \times W} \Big[ P_i \log Q_i \\
+ (1 - P_i) \log (1 - Q_i) \Big]
\end{multline}

This composite loss ensures both accurate per-pixel predictions and global shape alignment.

Note that to ensure comparability across losses, we apply a scaling coefficient to each attention loss term so that their magnitudes are roughly aligned; this coefficient can be treated as a tunable hyperparameter.

\section{Implementation details.}
\label{appendix:implementation_details}
\subsection{Attention Map Extraction Layer Justification} 

\begin{table}[t]
\centering
\setlength{\tabcolsep}{16pt}

\begin{tabular}{lc}
\toprule
\textbf{Layers used} & \textbf{Accuracy} $\uparrow$ \\
\midrule
All layers & 57.27\% \\
First 10 layers & \textbf{63.77\%} \\
Last 10 layers & 60.20\% \\
\bottomrule
\end{tabular}
\caption{Effect of layer selection in TinyLLaVA. Consistent with \cite{zhang2024redundancy}, focusing on early layers yields the highest accuracy, indicating that information critical for attention map extraction is concentrated in the initial stages.}
\label{tab:layer_ablation_tinyllava}
\end{table}

We extract attention maps capturing the alignment between visual and textual modalities from three representative LVLM, TinyLLaVA \citep{zhou2024tinyllavaframeworksmallscalelarge}, InternVL2 \citep{chen2024internvl}, and ChartGemma \citep{masry2024chartgemmavisualinstructiontuningchart}, which serve as the basis for gaze-guided attention refinement. 

Following the work of \cite{zhang2024redundancy}, which identified that the initial layers of LVLMs like LLaVA play a key role in cross-modal interaction, we conducted an analysis on TinyLLaVA to determine which layers were most effective for our task. As seen in Table~\ref{tab:layer_ablation_tinyllava},  our findings showed that focusing on the first 10 layers resulted in an accuracy of 63.77\%, a significant improvement over using all layers (57.27\%) or only the last 10 layers (60.195\%). This demonstrates that the information crucial for attention map extraction is primarily concentrated in the model's earlier layers.

We adopt a similar strategy for InternVL2. Due to its deeper architecture and higher-resolution visual encoding, we extracted attention from twelve layers (instead of ten) for both the 4B and 8B versions. For ChartGemma, we aggregated attention maps from the first six layers, as it is shallower than both TinyLLaVA and InternVL2.

\subsection{Training Details} We finetuned TinyLLaVa  using Low-Rank Adaptation (LoRA) on 15 epochs with learning rate of $1\times 10^{-4}$ with batch size 4, gradient accumulation steps 8, LoRA rank of 32 and LoRA $\alpha$ of 64. Moreover, we used RTX 3090 GPU with 24G VRAM. The fine-tuning was fast and took approximately 4 hours only.
% On average each iteration took 8.89 seconds and overall training time of more than 4 hours and 15 minutes. 
For ChartGemma, since the model has already instruction tuned on charts, we finetuned using LoRA on 7 epochs with learning rate of $5\times10^{-5} $, and batch size of 1 due to memory constraints with the same LoRA rank and $\alpha$ and with early stopping. For ChartGemma we used A100 with 40G VRAM. 
For InternVL2, we performed LoRA fine-tuning using 2× A100 GPUs with 80GB VRAM each. We used a learning rate of $5\times 10^{-5}$, batch size 2, gradient accumulation steps 16, LoRA rank of 32, and LoRA $\alpha$ of 64.

\section{Correlation Between Attention Interpretability and Accuracy}
\label{appendix:cc_accuracy}
\begin{figure}[h]
    \centering
    \includegraphics[width=0.85\linewidth]{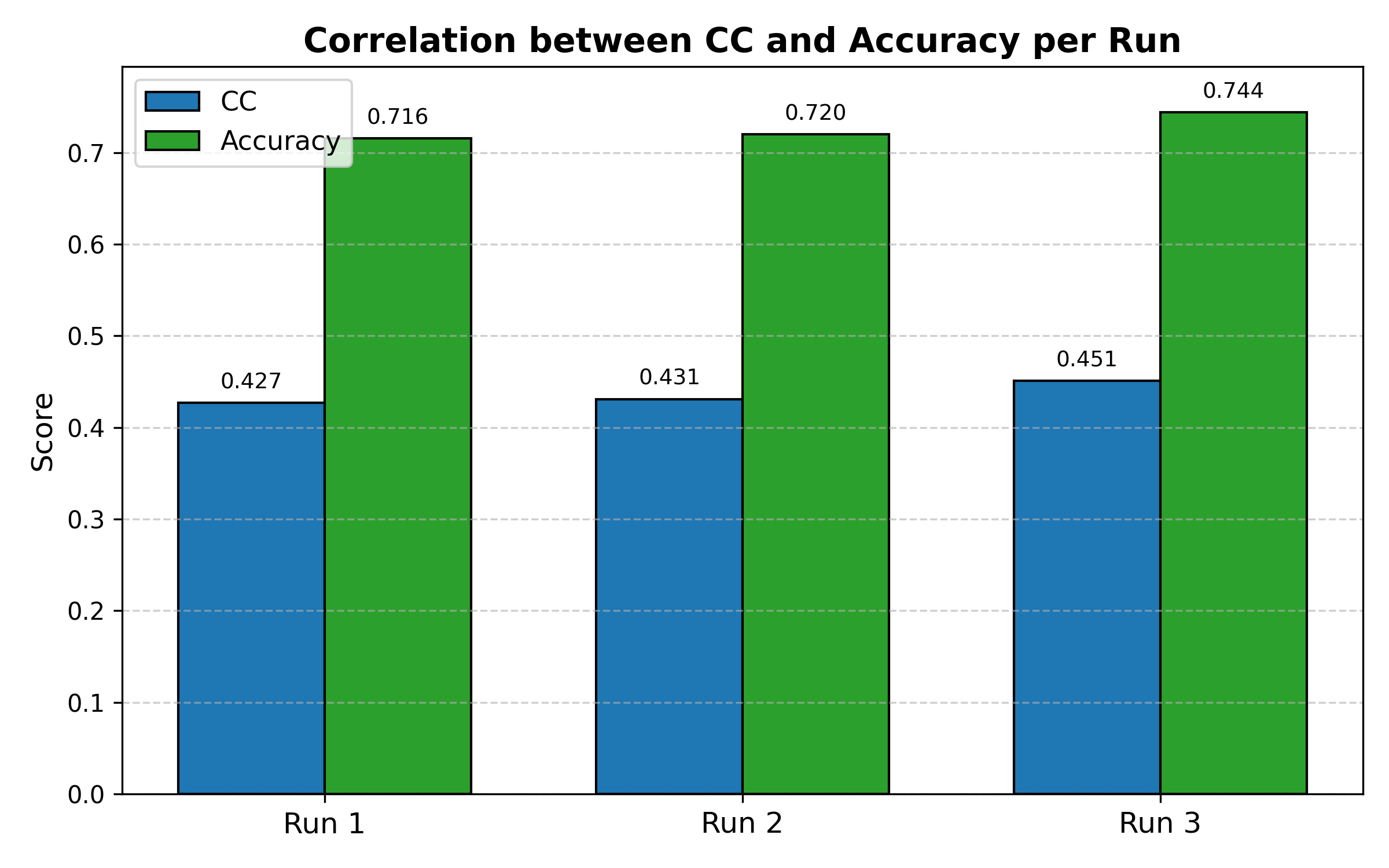}
    \caption{Correlation between CC (Correlation Coefficient) and accuracy across three independent runs of ChartGemma. Higher CC is consistently associated with higher QA accuracy, suggesting that more interpretable attention benefits task performance.}
    \label{fig:cc_vs_accuracy_runs}
\end{figure}

\begin{figure}[h]
    \centering
    \includegraphics[width=0.85\linewidth]{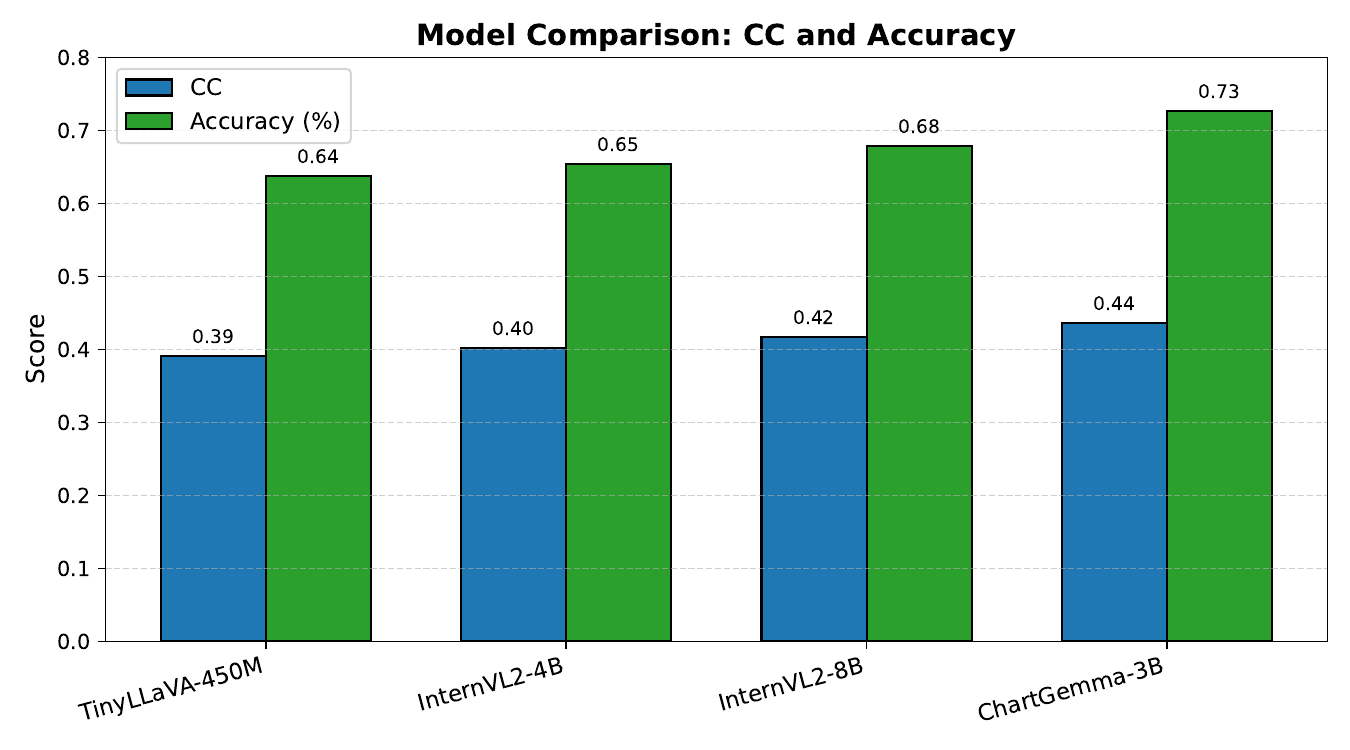}
    \caption{Model-level comparison of CC and QA accuracy across four models. Models with higher CC (e.g., InternVL2 and TinyLLaVA) tend to achieve higher QA accuracy, reinforcing the link between attention alignment and task effectiveness.}
    \label{fig:cc_vs_accuracy_models}
\end{figure}

As can be seen in Figures \ref{fig:cc_vs_accuracy_runs} and \ref{fig:cc_vs_accuracy_models}, there is a positive correlation between attention interpretability and accuracy.

\vspace{1em}

\begin{figure*}[!t]
  \centering
  \includegraphics[width=\textwidth]{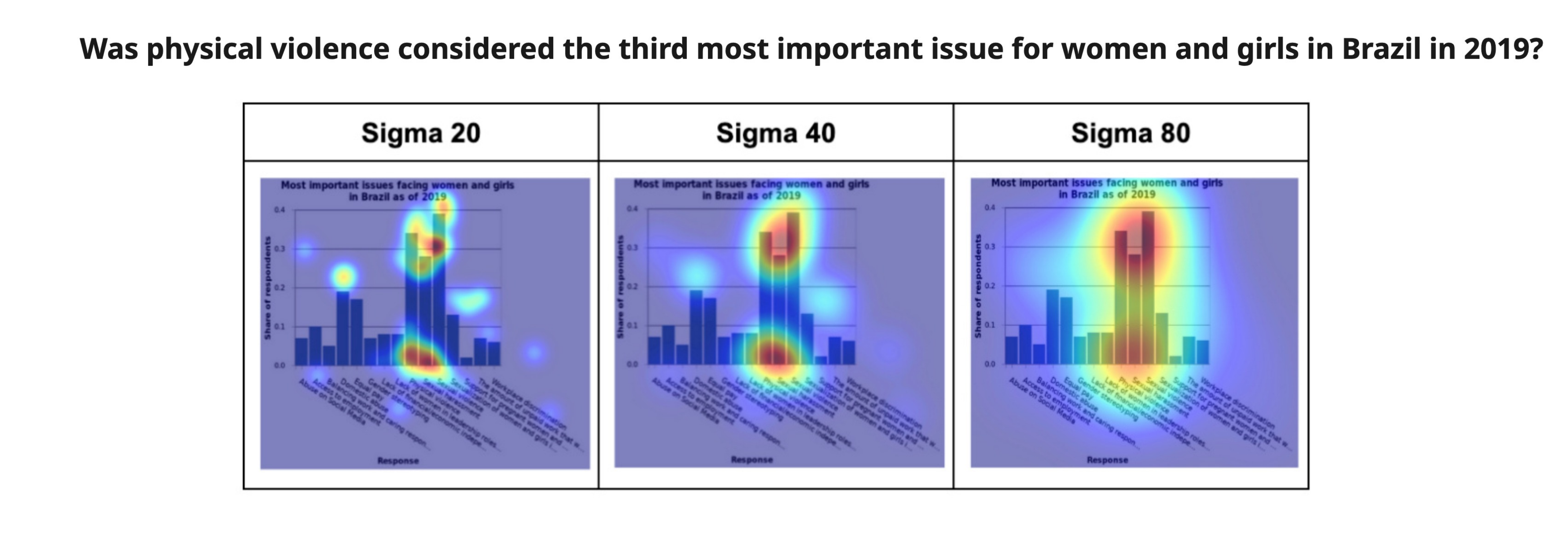}
  \caption{Comparison of different $\sigma$ values and its effect on the human gaze map.}
  \label{fig:sigma-choice}
\end{figure*}

\section{Generalization}
\label {appendix:generalization}

\begin{table}[t]
\centering
\setlength{\tabcolsep}{4pt}
\resizebox{\linewidth}{!}{%
\begin{tabular}{lcc}
\toprule
\textbf{Model} & \textbf{Relaxed Accuracy} \\
\midrule
\textbf{No Fine-tuning} \\
\quad InternVL2-4B (Pre-trained) & 81.52 \\
\quad InternVL2-8B (Pre-trained) & 83.28 \\
\addlinespace
\midrule
\textbf{Fine-tuned on ChartGaze, lang loss only} \\
\quad InternVL2-4B & 49.15 \\
\quad InternVL2-8B & 50.28 \\
\addlinespace
\midrule
\textbf{Fine-tuned on ChartGaze, lang+attention loss} \\
\quad InternVL2-4B & 51.42 \\
\quad InternVL2-8B & 53.08 \\
\bottomrule
\end{tabular}
}
\caption{Model performance on the ChartQA test set after fine-tuning with different strategies on our dataset.}
\label{tab:chartqa_performance1}
\end{table}

\begin{table}[t]
\centering
\setlength{\tabcolsep}{4pt}
\resizebox{\linewidth}{!}{%
\begin{tabular}{lcc}
\toprule
\textbf{Model} & \textbf{Relaxed Accuracy} \\
\midrule
\textbf{No Fine-tuning} \\
\quad InternVL2-4B (Pre-trained) & 81.52 \\
\quad InternVL2-8B (Pre-trained) & 83.28 \\
\addlinespace
\midrule
\textbf{Fine-tuned on ChartGaze, lang loss only} \\
\quad InternVL2-4B & 76.40 \\
\quad InternVL2-8B & 77.25 \\
\addlinespace
\midrule
\textbf{Fine-tuned on ChartGaze, lang+attention loss} \\
\quad InternVL2-4B & 78.64 \\
\quad InternVL2-8B & 79.85 \\
\bottomrule
\end{tabular}
}
\caption{Model performance on the ChartQA test set after fine-tuning with different strategies on our modified dataset.}
\label{tab:chartqa_performance2}
\end{table}

We conducted preliminary experiments on 150 open-ended ChartQA examples and observed an improvement of about 1\% in relaxed accuracy. This indicates that the method may generalize beyond Yes/No tasks, though further study is required to assess cost-effectiveness and scalability. We also evaluated InternVL2 models finetuned on our dataset using the ChartQA test set. As shown in Table \ref{tab:chartqa_performance1}, accuracy drops significantly because the models largely lose their language ability when trained only on Yes/No questions. Nevertheless, models trained with our loss still achieve better test performance. To further mitigate this issue, we re-trained the models on a version of our dataset where answers included full sentences (generated by GPT-4o from the question and Yes/No response) and then tested them on ChartQA. As shown in Table \ref{tab:chartqa_performance2}, this strategy reduces the performance drop, indicating that language ability is preserved to a greater extent. Once again, our trained model outperforms the baseline, suggesting that our attention refinement method yields models with stronger generalizability.

\end{document}